\documentclass[letterpaper, 10 pt, conference]{ieeeconf}  

\IEEEoverridecommandlockouts                              

\usepackage{cite}
\usepackage{amsmath,amssymb,amsfonts}
\usepackage{algorithmic}
\usepackage{graphicx}
\usepackage{textcomp}
\usepackage{lipsum}  
\usepackage{xcolor}
\usepackage[font=footnotesize]{caption}
\usepackage{subcaption}
\usepackage{gensymb}
\begin{document}

\title{\LARGE \bf A real-time, hardware agnostic framework for close-up branch reconstruction using RGB data\\
\thanks{This research is supported in part by USDA-NIFA through the Agriculture and Food Research Initiative, Agricultural Engineering Program (award No. 2020-67021-31958) and the AI Research Institutes program supported by NSF and USDA-NIFA under the AI Institute: Agricultural AI for Transforming Workforce and Decision Support (AgAID) (award No. 2021-67021-35344).}
\thanks{$^1$Collaborative Robotics and Intelligent Systems (CoRIS) Institute, Oregon State University, Corvallis OR 97331, USA {\tt\footnotesize \{youa, mehtaaru, strohbel, joseph.davidson, cindy.grimm\}@oregonstate.edu}}%
\thanks{$^2$Wageningen University \& Research, 6708 PB Wageningen, The Netherlands {\tt\footnotesize \{jochen.hemming\}@wur.nl}}
\author{Alexander You$^1$, Aarushi Mehta$^1$, Luke Strohbehn$^1$, Jochen Hemming$^2$, Joseph R. Davidson$^1$, Cindy Grimm$^1$}
}%

\maketitle

\begin{abstract}
Creating accurate 3D models of tree topology is an important task for tree pruning. The 3D model is used to decide which branches to prune and then to localize the pruning cuts with respect to the robot. Previous methods for creating 3D tree models have typically relied on point clouds, which are often computationally expensive to process and can suffer from missing data, especially with thin branches. In this paper, we propose a method for actively scanning along a primary tree branch, detecting secondary branches and reconstructing their 3D geometry using just an RGB camera mounted on a robot arm. We experimentally validate in real world trials that our setup is able to reconstruct primary branch center lines and radii within 2cm and 0.5cm, with a secondary branch miss rate of 25\% in lab and 45\% in outdoor conditions. Our framework is real-time and can scan at up to 4 cm/s with no loss in model accuracy or ability to detect secondary branches.

\end{abstract}

\section{Introduction}

Robotic fruit tree pruning is an active area of research~\cite{zahid_technological_2021} motivated by rising production costs and increasing worker shortages. Over the past several years, a team at the AI Institute for Transforming Workforce and Decision Support (https://agaid.org/) has been developing methods for robotic pruning. In previous field trials~\cite{you2023semiautonomous}, we demonstrated an integrated system capable of scanning a tree, searching for branches to prune, and then executing precision cuts using a combination of visual servoing and admittance control. In contrast with other existing pruning systems that use stereo vision for perception, our system relied solely on an RGB camera. However, this system had two notable deficiencies. First, it assumed the branches lay in a fixed (known) plane and (while adjusting up-down and left-right using visual servoing) simply moved the cutter towards the plane until contact was made, resulting in wasted execution time. Second, the system used a naive lawn-mower scanning pattern to find the branches, which resulted in a lot of wasted time. Follow-on work~\cite{parayil2023follow} addressed the second deficiency by actively searching for, and following, primary branches. Again, the assumption was made that the primary branch was in a known plane, and only the primary branch was reconstructed. In this paper we combine and extend these two approaches, removing the planar assumption by using point-to-point tracking to reconstruct the primary branch centerline in 3D. At the same time, we reconstruct the secondary branches attached to the primary one.

Historically, camera-based systems have used stereo in order to create a 3D point cloud of the tree, which is then post-processed into a skeleton or mesh. Typically, a large portion of the tree is scanned, either by setting the imaging system back from the tree or by using the robot to move the imaging system in a scanning pattern. Aside from computation time, the 3D point cloud must be aligned with the robot when it is then moved close enough to execute physical cuts. We argue that, for many pruning or measurement tasks, we can scan and cut as we go, reconstructing a much lighter-weight geometric model (centerline plus radius).

\begin{figure}[bt]
	\centering
\includegraphics[width = \columnwidth]{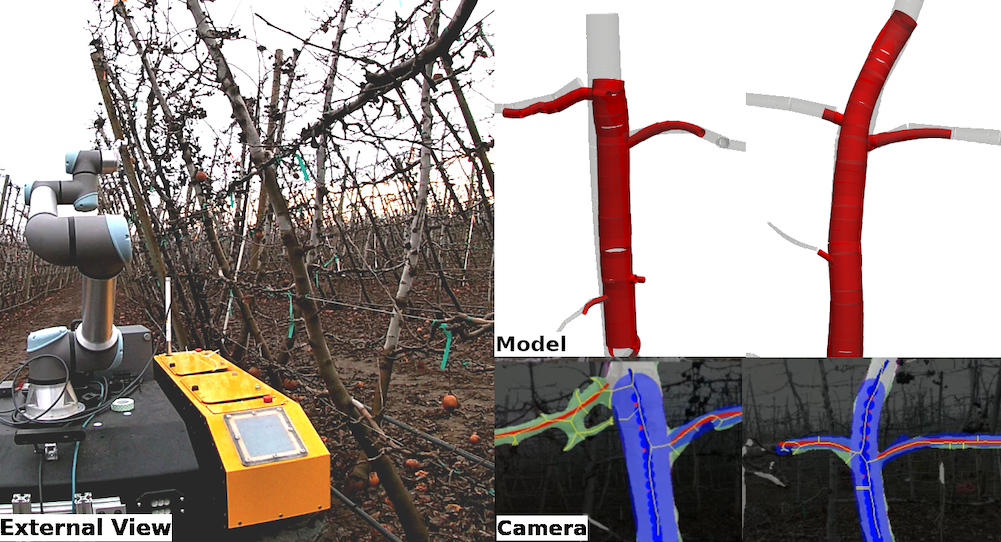}\caption{Our framework uses 2D RGB data and knowledge of the robot's kinematics, camera intrinsics, and hand-eye calibration to follow along a primary branch and create highly accurate 3D branch reconstructions of the primary and secondary branches that can be used immediately for picking a cut and executing it.}\label{fig:abstract}
    \vspace{-2ex}
\end{figure}

\begin{figure*}[bt]
	\centering
\includegraphics[width = 0.95\textwidth]{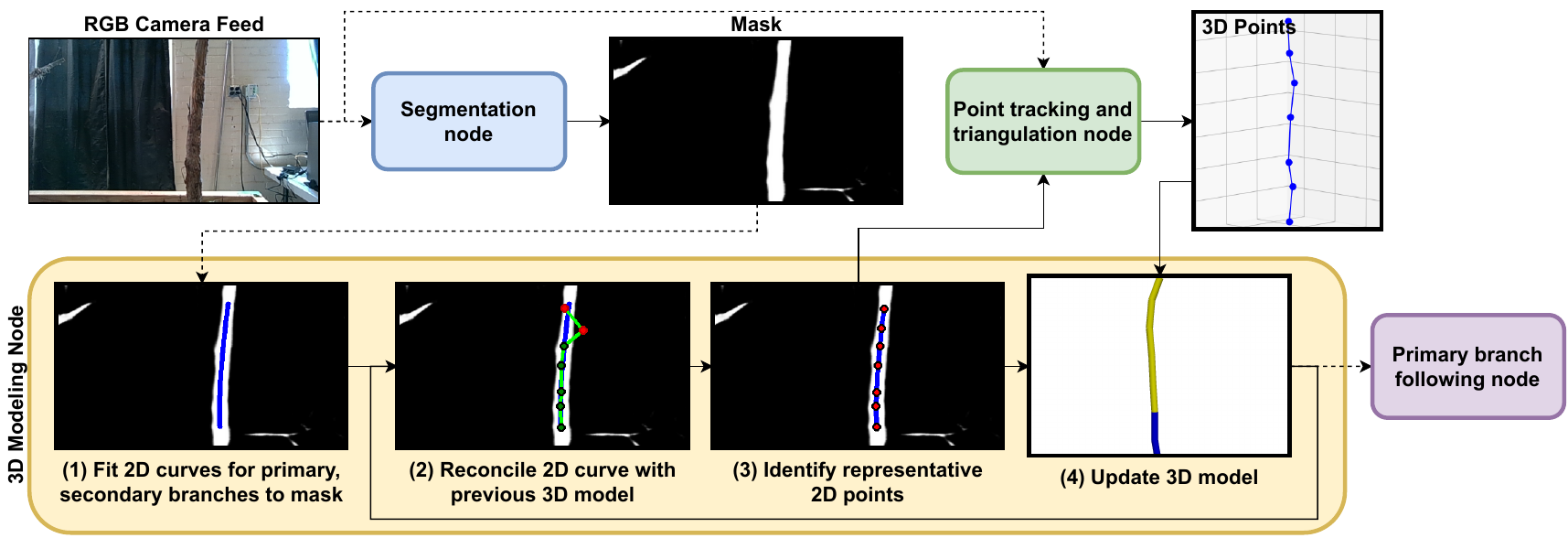}\caption{A system overview of our framework for branch scanning and reconstruction. It consists of 4 system nodes: A segmentation node producing binary masks; a point tracking plus triangulation node that reads in RGB images and returns 3D estimates for queried 2D pixels in an image; the core node that performs the branch reconstruction; and the controller node that moves the robot along the 3D primary branch. Dotted lines represent asynchronous system inputs.}\label{fig:system-diagram}
    \vspace{-2ex}
\end{figure*}

Our framework (Fig.~\ref{fig:abstract}) is designed to 1) reduce scanning time by just scanning along the primary branch; 2) reconstruct a lightweight 3D model ``on the fly''. Our system is capable of actively scanning along a primary branch, detecting secondary branches coming off of that primary branch, and creating 3D branch models that are accurate enough to be used for on-the-fly pruning decisions. The three innovative components of the system are 1) the branch-following controller, 2) using minimal computation to ``lift'' a 2D skeletonization to 3D, and 3) extending the existing 3D model with new data. To our knowledge, our work is the first to focus on real-time modeling for close-up tree views where only a portion of the tree is visible. One benefit of this approach is that it uses an \textit{arbitrary} off-the-shelf RGB camera mounted in an eye-in-hand configuration --- not specialized stereo or RGB-D cameras. 

We evaluate our framework in a simulated, lab, and orchard environment. In the lab the 3D reconstruction is remarkably accurate, with an average primary branch position accuracy of 4mm, radius error of 1.5mm and a branch detection rate of $81\%$. Outdoors, on a V-trellis apple orchard, we observed an average primary branch position accuracy of 19.4mm, radius error of 4.4mm and a branch detection rate of $47\%$. Furthermore, we can run at speeds of at least 4 cm/s without any noticeable loss of accuracy. 

A note on terminology: We refer to the primary branch as the one we are following while the secondary branches are those that grow out of the primary one. Depending on the tree architecture, this could be a leader and fruiting spurs (upright fruit offshoot (UFO) architecture), a trunk and main branches (tall spindle, V-trellis), or fruiting spurs off of a main branch (V-trellis). 

\section{Related Work}

The problem of creating a topological model of a tree (or any 3D object) is known as \textit{skeletonization}. We previously developed our own skeletonization algorithm~\cite{you2022semantics} that uses semantic knowledge about a specific tree architecture, plus a convolutional neural network measuring connectedness between two points, to model the quality of the reconstruction.  Other methods  exist for apple trees~\cite{karkee2015method,akbar2016novel,bucksch2008campino}, grapevines~\cite{botterill2017robot}, jujube trees~\cite{fu2023skeleton}, and rose bushes~\cite{cuevas2020segmentation}. All of these methods convert point clouds directly to skeletons, with the exception of~\cite{botterill2017robot} that instead builds a 3D model by matching canes across 2D images (similar to our own work). Also of note is the method of~\cite{cuevas2020segmentation} that adopts two key principles we also explore: segmentation masks to rectify 3D estimates and real-time algorithms. However, none of these methods actively control the camera to focus the reconstruction on the branches being scanned.

Our approach is inspired by structure from motion (SfM) techniques. These methods use a stream of 2D camera images with (potentially unknown) camera poses to create a 3D reconstruction of a scene by matching features between neighboring images and running an optimization method (bundle adjustment) to estimate the camera poses and construct the 3D scene. Such methods have been used for orchard-related applications such as fruit detection~\cite{gene2020fruit,liu2018robust}, though none seem to exist for modeling tree branches. Although full SfM algorithms are often computationally expensive and output high density point clouds, they can be scaled down to track just a few points, as we do here.

We also use ideas from active vision, i.e. using visual feedback to plan robot movement with the goal of maximizing information gain (detecting prunable branches). Our framework extends upon our previous Follow the Leader (FTL) controller~\cite{parayil2023follow}. This work only tracked the 2D projection of the primary branch in the image and did not create a 3D model of it. This meant that it was unable to keep a set distance from the camera to the branch if the branch did not lie in the estimated plane.  In the field of agri-food robotics, active vision has been used to localize greenhouse tomatoes~\cite{rapado2023development} and sweet peppers~\cite{barth2016design,lehnert20193d}.

\section{System Design}

Our work focuses on pruning in modern, high-density orchards. We assume that each tree architecture consists of \textit{primary} branches (vertical or horizontal) and \textit{secondary} side branches that grow perpendicularly off the primary branches. These side branches are candidates for pruning. Regardless of the specific fruit type (e.g. apple, sweet cherry, currant, etc.), training system (V-trellis, spindle, UFO), or specific pruning rules, all share a need to search for secondary branches, model their locations, and execute cuts. 

Our framework goals are to move the camera along the primary branch at a set distance of $z_{target}$, keep the branch centered in the camera's view, and identify secondary branches coming off the primary branch. The controller optionally uses {\em rotation} around the primary branch to disambiguate secondary branches pointing at the camera. The output is a 3D centerline and radius for each detected branch. For the secondary branches we reconstruct the skeleton from where it joins the primary branch to where it either ends or leaves the field of view of the camera.  

A flow chart of our system is shown in Figure~\ref{fig:system-diagram}. The core component of our framework is the system that constructs the 3D skeleton of the tree using a stream of binary masks as input, which are produced by our segmentation framework. The 3D modeling node combines 2D information from the mask with 3D information from the point tracking and triangulation node to iteratively build a model of the scanned tree. The movement of the robot is handled by the primary branch following controller node that uses the existing 3D model to follow along the branch at a set distance.

Our segmentation system is described in detail in~\cite{you2022optical}. It produces masks by augmenting RGB images with optical flow and processing them through a generative adversarial network, allowing for robust segmentation across various environments and lighting conditions. It should be noted that this segmentation system was trained entirely in simulation and with planar tree architectures that were similar, but not identical, to the lab and orchard experiments. For our real-world trials we also experimented with a YOLO-based RGB-only segmentation trained on manually labeled tree data, but it was largely unable to remove background trunks.

Although improving the segmentation would improve our results, our primary focus for this paper is the three novel components that allow for 3D perception of the scene as well as discovery of occluded branches.

\section{3D Branch Reconstruction}

In this section, we discuss the subsystem that iteratively constructs the 3D model of the primary and secondary branches. Each branch model consists of a linear sequence of 3D points (the centerline), as well as radius estimates at each point. The main steps are shown in the 3D Modeling Node box in Figure~\ref{fig:system-diagram}. At each iteration (which occurs after the robot has moved 10 mm from the last successful iteration), the modeling node takes in a segmentation mask and attempts to fit 2D B\'{e}zier curves for the primary and secondary branches and estimate their 2D radii (Figure~\ref{fig:system-diagram},~(1)); this operation is discussed below in Section~\ref{sec:curvefitting2d}. We then attempt to correct the existing 3D primary branch model by projecting it into the current image frame and checking if the projected points are consistent with the detected primary branch model (Figure~\ref{fig:system-diagram},~(2)), i.e. if at least 60\% of the projected branch points are in the foreground mask and within 4 px of a 2D curve fit to the centerline of the mask. There are three possible outcomes of the consistency-checking operation:

\begin{itemize}
    \item (Consistent) There is sufficient overlap. Extend and re-fit the 3D centerline with the new mask (see next paragraph).
    \item (Inconsistent) The current 3D curve does not project into the mask; we assume the mask data is bad and skip this frame.
    \item (Restart) If this is the 3rd inconsistent mask in a row we assume the current 3D model is wrong and delete all of the 3D curve visible in the last three frames. Re-initialize and continue.
\end{itemize}

Our goal for the 3D centerline is a set of 3D points spaced roughly 30 pixels apart (when projected back onto the image), each with a 3D radius estimate. We have two sources of 3D points; i) the 3D points from the current 3D centerline that project into the current image and ii) the 2D points of the mask centerline --- sampled with a spacing of 30 pixels --- (section~\ref{sec:curvefitting2d}) ``lifted'' into 3D using point triangulation (Section~\ref{sec:pixel3d}). We assume that the currently visible portion of the centerline can be modeled using a cubic B\'{e}zier curve. 

We use two methods to remove potential outliers from our two sets of 3D points, reprojection error and outlier detection using RANSAC when fitting the 3D B\'{e}zier curve. For 3D points from the current centerline, if the point does not project within 4 pixels of the 2D centerline, it is removed. For the lifted 2D points, we discard any points that have a depth value above $1m$ or a reprojection error of 4 pixels in the frame. For the RANSAC fit, we use an inlier threshold of $3cm$. Outlier points are then removed. If the number of inlier points falls below 60\% the frame is marked Inconsistent. 

For each kept point we estimate the 3D radius by $zr_{px} / f$, where $z$ is the camera frame's $z$-coordinate of the point, $r_{px}$ is the point's corresponding 2D pixel radius estimate, and $f$ is the camera focal length.

If this is a Restart, then we remove any current 3D points in the frame that project into the image and just use the 2D lifted points.


We follow the same procedure for finding the 3D centerline and radii of secondary branches. For newly detected secondary branches we perform an additional check to see if the branch is actually connected (and not a background branch). This check constructs a ray from the start of the  B\'{e}zier curve and determines if the ray comes within 3 cm of the primary branch's centerline. 

\subsection{2D curve fitting}
\label{sec:curvefitting2d}

\begin{figure}[bt]
	\centering
\includegraphics[width = 0.9\columnwidth]{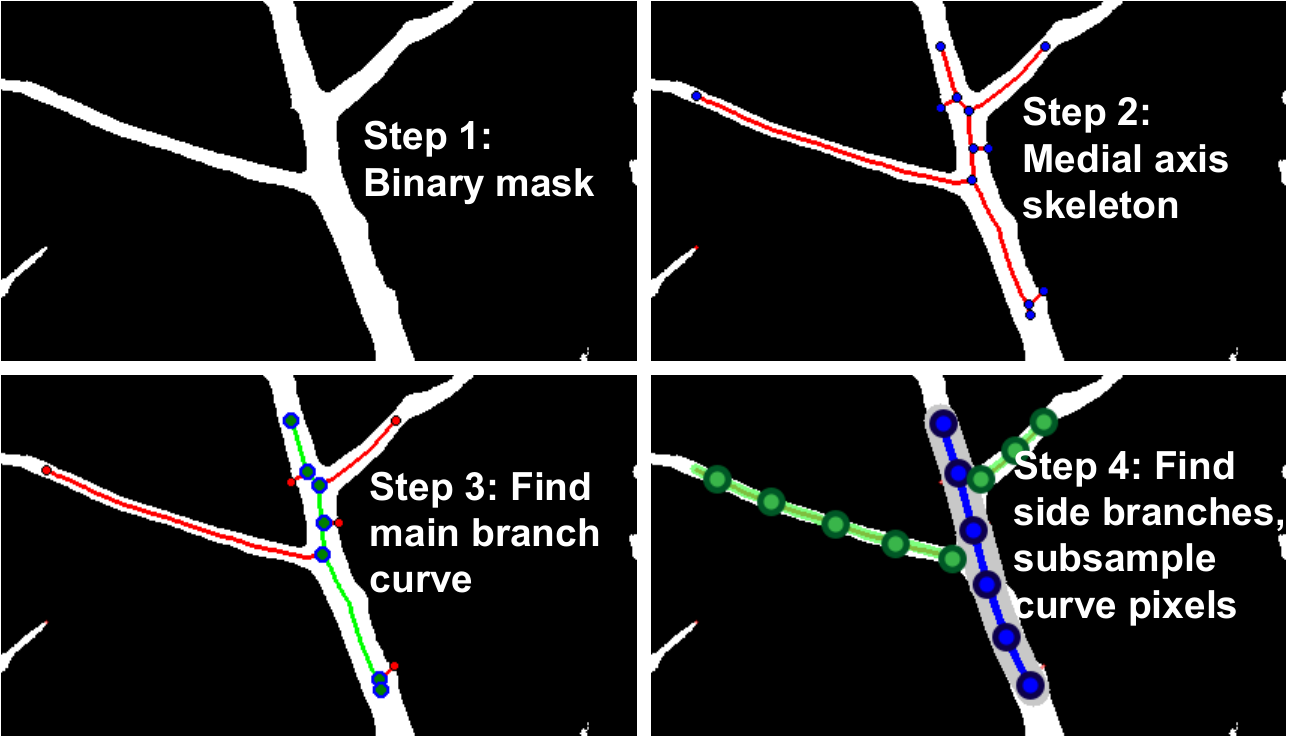}
\caption{The process for extracting 2D primary and secondary branches from a skeleton. We start with the medial axis skeleton and find the subpath with the best fit to be the main branch. We then check skeletal paths extending off the primary branch to see if they are sufficiently long and a B\'{e}zier curve can be fit to them. We subsample the curves to identify pixels for which we will estimate 3D positions.}\label{fig:skel}
    \vspace{-2ex}
\end{figure}

Figure~\ref{fig:skel} shows the process of locating the primary and secondary branch curves in a 2D mask. We start by using medial axis skeletonization~\cite{blum1967transformation} to obtain a pixel skeleton and radius estimates for each skeletal pixel. The pixel skeleton is then turned into a graph in which the nodes are all pixels at an endpoint (degree 1) or at a junction (degree 3 or above), with an edge between two nodes if they are connected by a linear segment of the skeleton (Step 2). These edges are oriented to point in the up ($-Y$) direction of the image, resulting in a multi-tree directed graph with no cycles.

To fit the primary B\'{e}zier curve, we run an exhaustive search of all possible pixel paths in the multi-tree (Step 3). For each pixel path, we perform a least squares regression to fit a cubic B\'{e}zier curve to the set of pixel points and select all inlier skeleton pixels that are within a given threshold (4 px) of the B\'{e}zier curve. The score of the pixel path is the size of the set of unique $y$-values of all inlier pixels. Once we have identified the highest scoring pixel path and associated B\'{e}zier curve $B(t)$, we subsample the curve to obtain representative 2D pixels (Step 4, blue curve) and match each subsampled pixel to the closest skeletal point to obtain its radius.

Once we identify the primary 2D branch curve, we then identify all secondary branch curves. First, from the original (undirected) pixel skeleton graph, we identify all edges that are connected to the inside of the identified primary branch; these edges represent the starts of potential subpaths for secondary branches. For each edge, we run an exhaustive search like before to find the subpath with the best fitting B\'{e}zier curve. If the curve pixel length and the inlier rate is high enough, we add the detection to the set of ``potential'' secondary branches.

\subsection{3D triangulation via pixel tracking}
\label{sec:pixel3d}

\begin{figure}[bt]
	\centering
\includegraphics[width = 0.9\columnwidth]{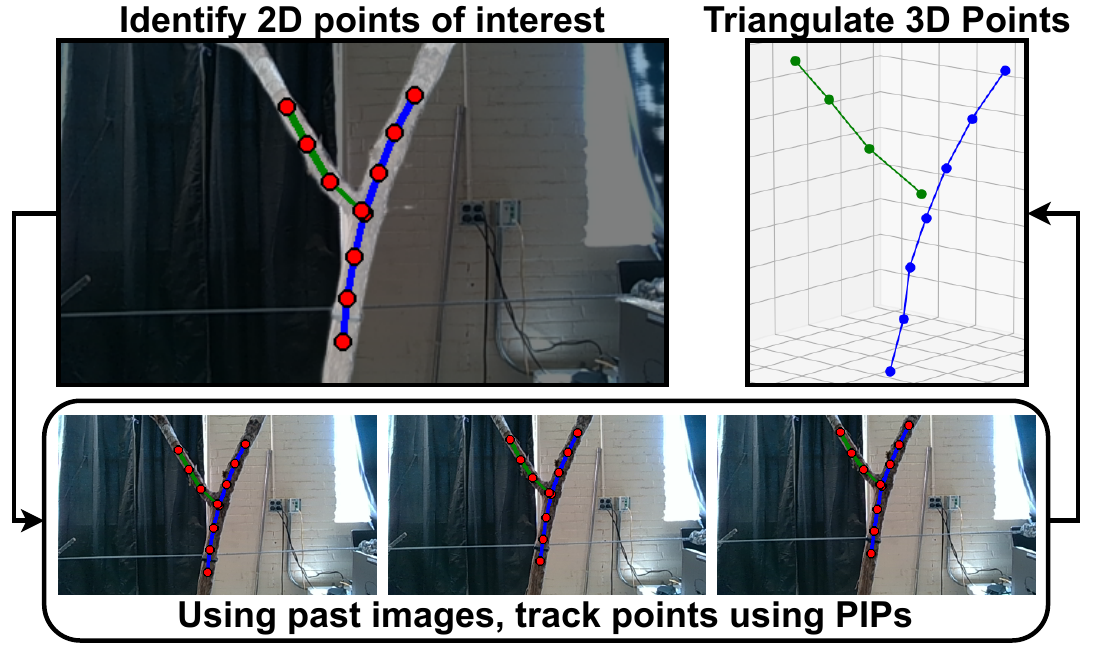}\caption{Using the Persistent Independent Particles point tracker, along with knowledge of the camera's poses and intrinsics, we can obtain 3D estimates of an arbitrary subset of pixels in an image.}\label{fig:pips}
    \vspace{-2ex}
\end{figure}

To obtain 3D point estimates from the identified 2D pixels, we use the Persistent Independent Particles (PIPs) method~\cite{harley2022particle}. This approach tracks a set of queried pixels over $M=8$ video frames, which are accumulated as the robot moves every 1.0 mm (Figure~\ref{fig:pips}). We then combine each pixel's trajectory with knowledge of the camera's world frame pose and intrinsics at each frame to triangulate each pixel's 3D coordinate.

Given a fixed camera calibration matrix $\mathbf{K} \in \mathbb{R}^{3\times3}$ and a series of $M=8$ camera poses defined by rotation matrices $\mathbf{R}_i \in SO(3)$ and translations $T_i \in \mathbb{R}^3$, we obtain a series of projection matrices $\mathbf{P}_i = \mathbf{K} \begin{bmatrix}\mathbf{R}_i & T_i \end{bmatrix} \in \mathbb{R}^{3\times4}$. 
Each projection matrix transforms homogeneous 3D points into 2D plus depth image coordinates. Denoting the rows of $\mathbf{P}_i$ as $\mathbf{p}_i^1$, $\mathbf{p}_i^2$, and $\mathbf{p}_i^3$, and given the list of $M$ pixel correspondences  $(u, v)_{j,i}$ ($j$ keypoints for each of $i=M$ camera frames), we can define the matrix $\mathbf{D}_j \in \mathbb{R}^{2M\times4}$ as follows:

\begin{equation}
    \mathbf{D}_j = \begin{bmatrix}
         \mathbf{D}_{j,1} \\ \vdots \\ \mathbf{D}_{j,M}
    \end{bmatrix}, \mbox{ where }
    \mathbf{D}_{j,i} = \begin{bmatrix}
        \left(u_{j,i}\mathbf{p}_i^3\right) - \mathbf{p}_i^1 \\
        \left(v_{j,i}\mathbf{p}_i^3\right) - \mathbf{p}_i^2
    \end{bmatrix}
\end{equation}


The problem of finding a point $\mathbf{X}_j \in \mathbb{R}^3$ that minimizes the reconstruction error across all views is the same as finding a point $\mathbf{\underline{X}}_j \in \mathbb{R}^4, \Vert\mathbf{\underline{X}}_j\Vert = 1$ that minimizes the algebraic least-squares error $\Vert\mathbf{D}_j\mathbf{\underline{X}}_j\Vert^2$, which can be solved with singular value decomposition (SVD).

\section{3D branch-following controller}

\begin{figure}[bt]
	\centering
\includegraphics[width = 0.8\columnwidth]{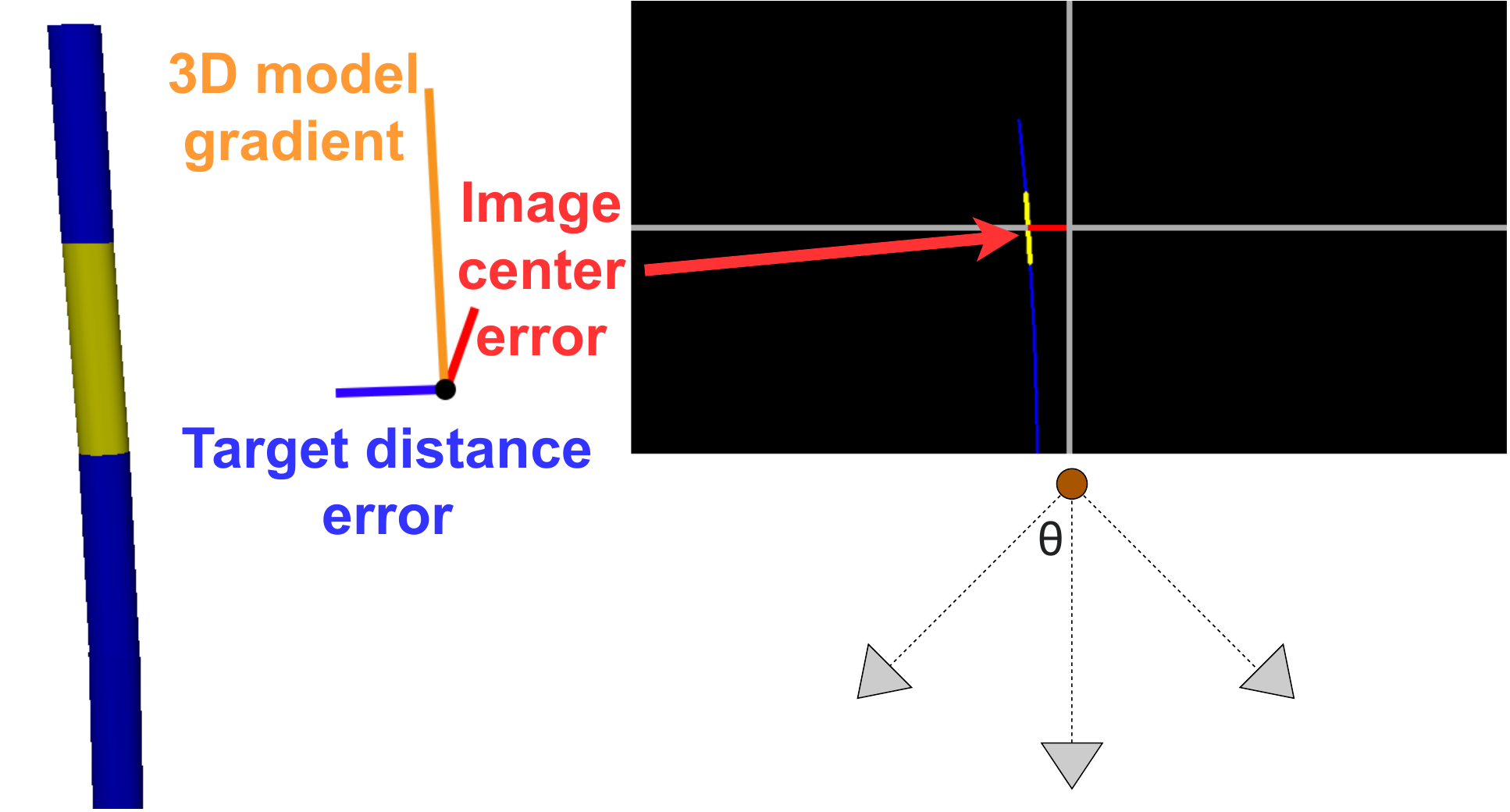}\caption{Components of the primary branch following controller. The controller follows the gradient of the 3D centerline that lies in front of the camera, with velocity offsets to keep a fixed distance from the branch and to keep the branch centered in the image (top-right). To gain a view of potentially occluded branches, it also periodically cycles through 3 camera orientations at an angle $\theta$ (bottom-right, shown from a top view).}\label{fig:controller}
    \vspace{-2ex}
\end{figure}

In this section we describe the controller used to follow along the primary branch (Figure~\ref{fig:controller}). This controller also performs the function of periodically rotating the camera around the primary branch, with the goal of gaining a view of branches that may be occluded by the original view.

\begin{figure*}[h]
	\centering
\includegraphics[width = 0.95\textwidth]{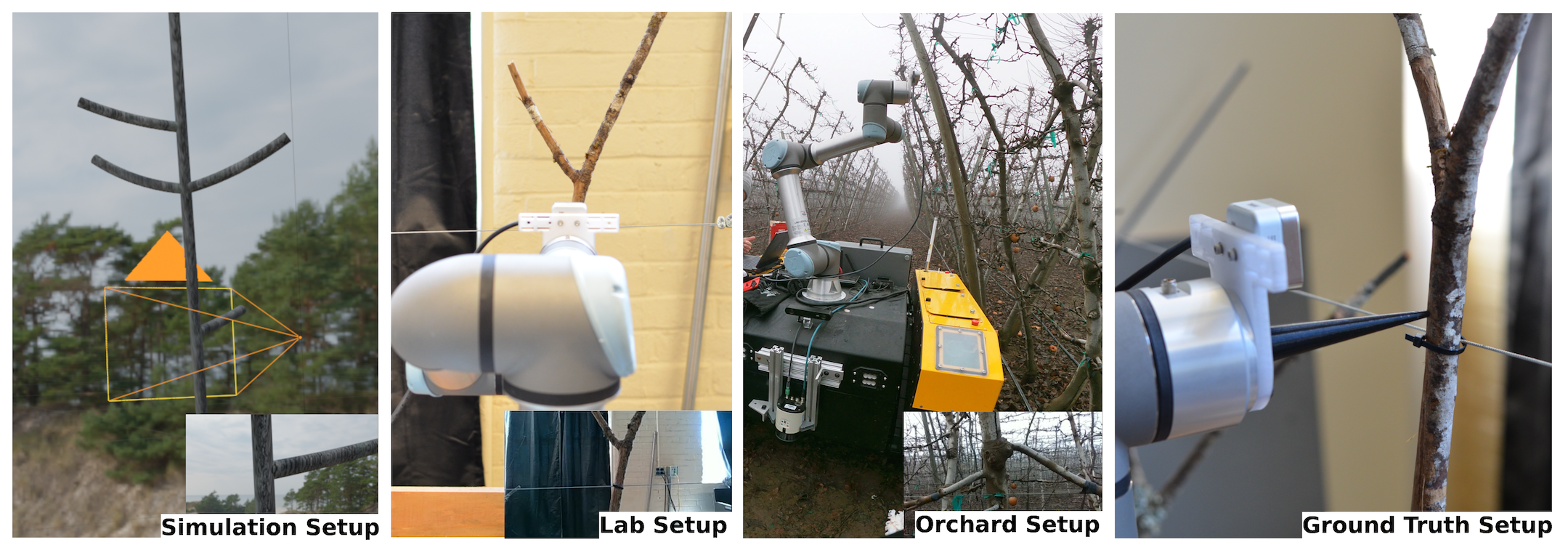}\caption{Setups used for evaluating our framework. Left To Right: Simulated Blender environment showing a mock spindle tree labelled with the camera frame and direction in orange (camera view inset), Lab experimental setup using branches (camera view inset), Orchard experimental setup (camera view inset), Ground truth setup using measurement probe}\label{fig:experimentscombined}
    \vspace{-2ex}
\end{figure*}

To output the camera-frame velocity that tracks the primary branch, we use the 3D primary branch centerline described in the previous section. Supposing that the camera has an image width of $w$ and height $h$, we project the 3D primary branch centerline into the camera image frame and identify the pixel that falls closest to the vertical center at $\frac{h}{2}$. Let the identified pixel be $(u,v)$ with the corresponding camera frame 3D point $p = (x, y, z)$ having a gradient of $\nabla$. The formula for the velocity is a simple proportional controller with scaling factors $k_u$ for the horizontal pixel difference and $k_z$ for the target distance error (this velocity is subsequently scaled to a predetermined speed):

\begin{equation}
    v_{tool} = \nabla + k_u \left(u - \frac{w}{2}\right) + k_z \left( z - z_{target} \right)
\end{equation}

In addition, to gain a view of branches that are either directly in front of or behind the primary branch, the controller can also be configured to periodically rotate the camera by a given angle $\theta$ in the world $Z$-plane about $p$ after moving a set distance up in the world (Figure~\ref{fig:controller}, bottom-right). It is useful to characterize this distance as a \textit{frequency} with respect to the length of the camera's vertical field of view at a distance of $z_{target}$; ideally the frequency should be greater than 1 to make sure the camera will not accidentally skip potentially occluded branches.

\section{Experiments}

To validate the quality of the models produced by our framework, we performed experiments in simulation, in a lab environment on real tree branches, and on trees in a commercial orchard (Figure~\ref{fig:experimentscombined}). The simulation experiments were primarily used to compare different algorithm parameters (speed of travel, amount of rotation, and frequency of image sampling).

\subsection{Simulated Experiments}


The simulated experiment uses the Blender environment (Figure~\ref{fig:experimentscombined} left) to render a mock primary branch with secondary side branches. The primary branch is modeled as a randomized cubic B\'{e}zier curve extending from $z=0$ to $1$ meters in the world frame. Within the range of $z \in [0.325, 0.75]$, we add side branches by uniformly sampling $z$-values and placing branches with random curvature at random orientations. The branch elevations are limited to $[-15\degree, 45\degree]$, while the yaw can be any value.

The simulated robot is a UR5e manipulator (Universal Robots, Odense, Denmark) with a 320x240 camera coincident with the end effector frame. Each experiment started with sending the simulated robot to a home position at $z=0.275$ and running the framework until completion (the robot reaches $z=0.75)$ or failure (typically due to hitting a kinematic singularity).

To evaluate the ability of the controller to accurately detect branches and reconstruct their 3D geometry, we ran experiments with 5 different parameter sets for the branch-following controller. The baseline parameter set is no rotation in which the robot simply scans straight up the tree without changing its viewpoint. The other parameter sets add angular rotations of $\{22.5\degree, 45\degree\}$ with rotation frequencies of $\{1.5, 2.5\}$ per vertical FoV, which for our setup with a distance $z_{target} = 0.20$ was 0.218 m.

\subsection{Lab Experiments}

The setup for the laboratory experiments is shown in Figure~\ref{fig:experimentscombined} (middle). We use a UR5e with an Intel RealSense D405 camera mounted to the wrist flange. For each experiment, we secured a branch to the trellis and ran the system to extract a 3D model. To obtain ``ground truth'' data for the branch, we attached a probe to the robot and manually sampled the location of the surface of the primary and secondary branches. We also used digital calipers to record the radius at each sampled point.

The lab experiments follow the same procedure as the simulated ones. We ran our experiments for 4 different branches with 4 parameter sets: the baseline controller with no rotation at $\{0.02, 0.05, 0.10\}$ m/s, as well as the rotating controller with a lookat angle of $22.5\degree$ and 1.5 rotations per vertical FoV at $0.05$ m/s. 

\subsection{Orchard Experiments}

The setup for the orchard experiments is shown in Figure~\ref{fig:experimentscombined} (right). We used a Clearpath Warthog-mounted UR5e with an Intel RealSense D435 camera mounted to the wrist flange. The orchard featured a v-trellis setup of 'Envy' apple trees in the dormant season. The ground truth was determined as in the lab setup. 

The experiments followed a slightly modified procedure due to experimental limitations. Real trees are significantly larger and have noisier backgrounds than the proxy lab experiments and the (virtual) trees used to train the image segmentation algorithm. For this reason, we experimented with two different distances from the primary branch (0.4 and 0.6m). Instability in the 3d model prevented us from safely rotating the robot. We scanned 4 different trees, chosen at random. Each tree was scanned 5 times at each distance, for a total of 40 scans.


\section{Results and Discussion}
Table~\ref{table:results_combined} shows a summary of the experiments (simulated, lab and orchard), and Figure~\ref{fig:abstract} shows several high-quality sample reconstructions. 

\subsection{Simulated Experiments}

\tabcolsep=0.11cm
\begin{table*}[]
\centering
\begin{tabular}{|l|ccccc|cccc|cc|}
\hline
 & \multicolumn{5}{|c|}{Simulated Experiments} & \multicolumn{4}{|c|}{Lab Experiments}  & \multicolumn{2}{|c|}{Orchard Experiments} \\ \hline
Rot. Angle \degree         & 0                 & 22.5              & 45                & 22.5              & 45 & 22.5 & 0 & 0 & 0 & 0 & 0              \\
Rot. Hz              & 0                 & 1.5               & 1.5               & 2.5               & 2.5 & 1.5 & 0 & 0 & 0 & 0 & 0              \\
Speed (m/s) & N/A & N/A & N/A & N/A & N/A & 0.05 & 0.02 & 0.05 & 0.10 & 0.04 & 0.04 \\
Distance (m) & N/A & N/A & N/A & N/A & N/A & 0.2 & 0.2 & 0.2 & 0.2 & 0.4 & 0.6 \\
Trials & 60 & 60 & 60 & 60 & 60 & 12 & 12 & 12 & 12 & 20 & 21 \\
\hline

PB Resid. (mm)       & 3.5 (4.0)         & 4.2 (4.9)      & 4.5 (8.9)      & 3.8 (3.7)      & 4.8 (7.5) & 4.2 (2.2) & 11.8 (6.3) & 6.5 (5.1) & 5.0 (4.1) & 19.5 (19.0) & 21.8 (20.4)  \\
PB Rad. (mm)     & 0.5 (0.1)       &  0.4 (0.4)   & 0.4 (0.4)    & 0.4 (0.3)    & (0.5) (0.4) & 1.5 (1.0) & 1.8 (1.6) & 1.0 (1.4) & 1.7 (1.3) & 4.4 (4.0) & 4.0 (1.3)  \\ \hline
SB Resid. (mm) & 5.7 (9.3)       & 11.0 (23.7)      & 6.1 (10.0)       & 9.7 (23.6)       & 7.3 (15.2) & 9.6 (8.4) & 12.4 (20.7) & 8.9 (11.0) & 7.1 (4.4) & 41.7 (34.5) & 31.1 (25.3) \\
SB Rad.(mm)        & 3.8 (0.9) & 3.8 (1.0) & 3.9 (0.9) & 3.8 (1.0) & 3.9 (0.9) & 1.3 (1.2) & 1.5 (1.5) & 1.1 (1.1) & 1.4 (1.2) & 2.4 (2.4) & 2.4 (2.4)\\
Spurious \%          & 2.3            & 0.8            & 4.5            & 0.3            & 0.4 & 43.7 & 22.9 & 24.3 & 25.0    & 23.9 & 24.8       \\
Missed \%          & 7.0            & 3.4            & 7.9           & 4.5            & 8.5 & 18.7 & 18.7 & 20.8 & 25.0  & 45.7 & 59.7        \\
\hline
\end{tabular}
\caption{Results of the simulated and real experiments for each parameter set (Rot. = Rotation, Freq.PB = primary branch, SB = secondary branch, Resid. = residual error, Rad. = radius RMSE). Standard deviation values are given in parentheses.}
\label{table:results_combined}
\end{table*}

First, we focus on the results of the baseline controller with no rotation before analyzing the results of adding rotation. As a whole, the reconstruction accuracy of the framework in simulation was remarkably accurate, with an average primary branch residual of just 3.5mm and a radius mean square error 0f 0.5mm. These values were generally positively skewed, characterized by infrequent, but large, outliers. 

We only missed secondary branches 7\% of the time, even without rotation. Typically, the branches that were missed were either pointing straight towards or away from the camera. For the branches that were detected, position error was under 5.7mm and radius 3.8mm. Large errors tended to have two causes: Either a background point was selected for 3D estimation and not filtered out, or poor 3D estimates in the secondary branch caused the cubic B\'{e}zier curve to have an unusual shape with sharp bends.

\begin{figure}[bt]
	\centering
\includegraphics[width = \columnwidth]{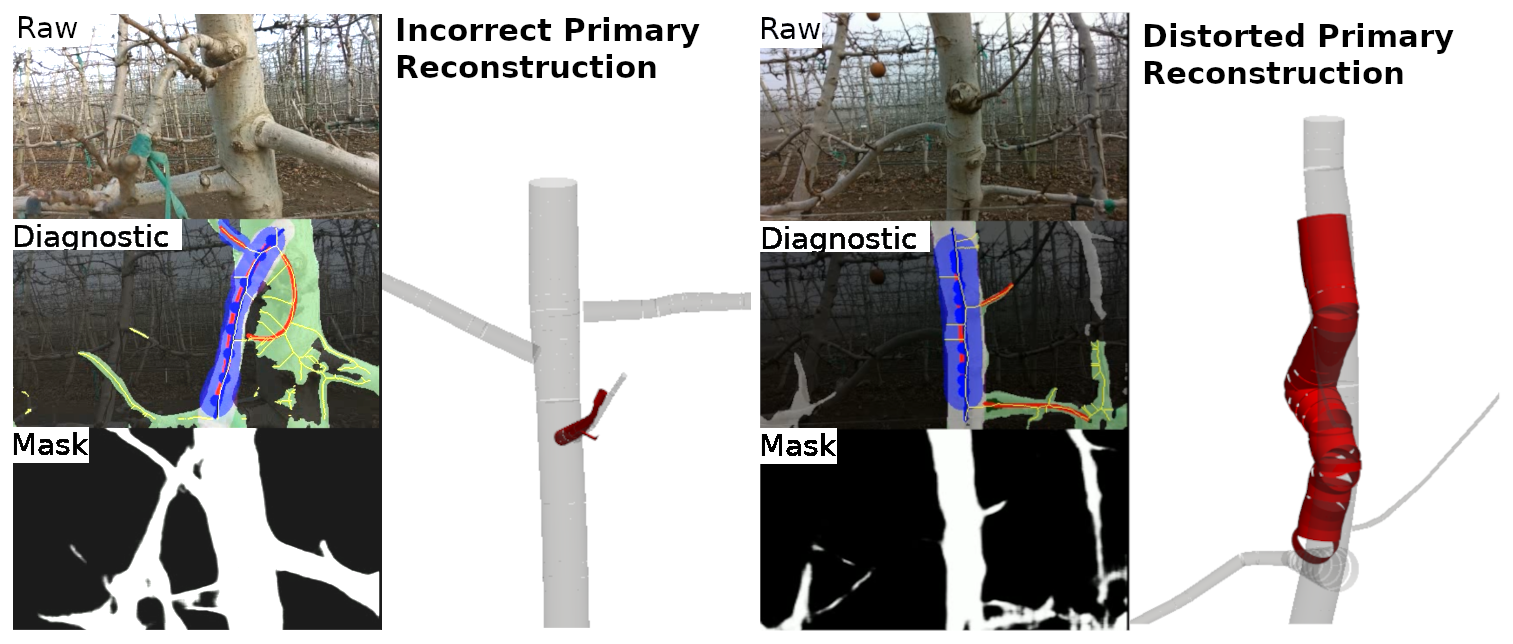}\caption{Examples of outlier reconstructions with diagnostic images. For each set: on the right is the reconstruction with the ground truth in gray and the model result in red. On the left are associated camera images, from top to bottom: the raw camera image, a diagnostic with segmentation of the primary branch in blue and secondary in green, and finally optical flow masks.}\label{fig:outliers}
    \vspace{-2ex}
\end{figure}

The effects of adding rotation to the controller were mixed. First, there were not any notable differences between the two rotation frequencies (1.5 and 2.5), indicating that 1.5 is preferred since the robot will spend less time rotating the camera. Regarding the different rotation angles, there was a decrease in branch miss rate using the 22.5 degree value, which is expected since the robot should be able to spot branches that initially point directly toward or away from the camera. This came with reduced accuracy. Surprisingly, the  detection advantage disappeared when increasing the rotation to 45 degrees and led to increased spurious detections.

One other notable weakness in secondary branch detection was redetecting the same branch multiple times. Rarely, the system would produce a completely spurious branch.


\subsection{Lab Experiments}

The lab experiments had higher primary residuals and radius errors. The main source of error was a model point incorrectly projecting into the background and never being removed. Unlike the simulator experiments, including rotation in the controller reduced the presence of large residuals caused by poor 3D estimates since the system could detect such errors as inconsistent with the mask and correct them (but did increase spurious detections).

For the secondary branches, the miss rate was higher (24\%) but the position errors were similar. The primary cause of the lower success rate was the real branches being thinner and irregularly shaped. The background wall was also closer to the robot than the background in the simulated environment, which caused the masks to be noisier. 

Interestingly, most metrics improved as speed increased. This is probably because the images were spaced further apart, which increased the baseline for reconstruction.

\subsection{Orchard Experiments}

In the orchard we needed to place the camera further from the primary branch because it was an order of magnitude bigger than the simulator or lab experiments. The segmentation model struggled with several challenges --- varying lighting, trellis architecture features such as struts, wires, threads etc, and a challenging background. We saw particularly large outliers in primary branch reconstruction, illustrated in Figure~\ref{fig:outliers} due to: (a) the complexity of the tree and its background resulting in the wrong primary branch being selected and (b) adjacent trees around 30-40cm behind resulting in the selection of background points distorting the primary. Side branches could not be connected to such primary branches in addition to any thin and irregular branches missed by the segmentation model. This resulted in a significantly higher miss rate than the lab experiments of 45.7\% when up close.

Primary branch residuals were in the order of 20mm and radius errors of 4mm. Side branch accuracy improved at the larger distance but this distance also resulted in a much worse detection rate. Secondary branch residual and radius errors were in the range of 35mm and 2mm. 

\subsection{Runtime}

We used a computer with an i7-11800H 8-core processor and an NVIDIA GeForce RTX 3060 Laptop GPU, running ROS2 Humble on Ubuntu 22.04. With this setup and 320x240 image inputs, the average runtimes to process one input for each node were:

\begin{itemize}
    \item Segmentation node: 0.037 s (26.9 Hz)
    \item 3D curve modeling node: 0.146 s (6.8 Hz). This runtime includes the synchronous call to the point tracking and triangulation node, which took 0.068 s (14.6 Hz)
    \item Primary branch following node: 0.00011 s (9300 Hz)
\end{itemize}

\noindent These runtimes are suitable for real-time use, with the main bottleneck (the 3D curve modeling node) still running nearly seven times per second. 


\subsection{Limitations}
The segmentation model is currently trained on a handful of tree architectures and camera distances and so fails fairly regularly. Training on the specific tree architecture to be scanned and including confounding artifacts, such as ties, wires and posts, would improve the mask generation. The method for determining if the mask is consistent with the 3D reconstruction is currently overly simplistic. A more rigorous approach, using a probabilistic particle model SLAM approach, could address this problem, at increased computational cost. The rotation is similarly simplistic, just happening at a fixed frequency instead of being driven by inconsistencies between the masks and the 3D model.

\section{Conclusion}

In this paper, we introduced a framework for scanning along a primary branch while constructing a model of the branch and the secondary branches attached to it. This framework is motivated by the need to quickly and efficiently locate prunable secondary branches and their potential cut points. By combining traditional 2D skeletonization methods with deep neural networks for segmentation and 3D point estimation, we developed a system that can create accurate branch models while running at practical speeds. This work demonstrates that is is possible to use active scanning with a light-weight 3D model and just an RGB camera to create on-the-fly models for making pruning decisions.

\bibliographystyle{./bibliography/IEEEtran}
\bibliography{./bibliography/pruning2022.bib}

\end{document}